\documentclass{article}

\usepackage{arxiv}

\usepackage[utf8]{inputenc} 
\usepackage[T1]{fontenc}    
\usepackage{hyperref}       
\usepackage{url}            
\usepackage{booktabs}       
\usepackage{amsfonts}       
\usepackage{nicefrac}       
\usepackage{microtype}      
\usepackage{lipsum}
\usepackage{graphicx}
\graphicspath{ {./images/} }

\usepackage[table]{xcolor}
\usepackage{subfig}
\usepackage{amsmath}

\usepackage{stackrel}
\usepackage{graphicx}
\usepackage{multirow}
\usepackage[export]{adjustbox}

\usepackage{pifont}

\title{ A novel Graph Transformer Framework for Gene Regulatory Network
Inference}

\author{
 Binon Teji \\
  Network Reconstruction \& Analysis (NETRA) Lab ,\\
Department of Computer Applications, \\ Sikkim
University, Sikkim, India, 737102  \\ \texttt{bteji.20pdca01@sikkimuniversity.ac.in} \\
   \And
 Swarup Roy \\
   Network Reconstruction \& Analysis (NETRA) Lab, \\
Department of Computer Applications, \\ Sikkim
University, Sikkim, India, 737102  \\ 
  \texttt{sroy01@cus.ac.in} \\
}

\date{}

\makeatletter
\renewcommand{\@date}{}
\makeatother

\begin{document}
\maketitle

\begin{abstract}
The inference of gene regulatory networks (GRNs) is a foundational stride towards deciphering the fundamentals of
complex biological systems. Inferring a possible regulatory link between two genes can be formulated as a link prediction problem.
Inference of GRNs via gene coexpression profiling data may not always reflect true biological interactions, as its susceptibility to noise
and misrepresenting true biological regulatory relationships. Most GRN inference methods face several challenges in the network
reconstruction phase. Therefore, it is important to encode gene expression values, leverege the prior knowledge gained from the
available inferred network structures and positional informations of the input network nodes towards inferring a better and more
confident GRN network reconstruction.
In this paper, we explore the integration of multiple inferred networks to enhance the inference of Gene Regulatory Networks (GRNs).
Primarily, we employ autoencoder-based embeddings to capture gene expression patterns directly from raw data, preserving intricate
biological signals. Then, we embed the prior knowledge from GRN structures transforming them into a text-like representation using
random walks, which are then encoded with a masked language model, BERT, to generate global embeddings for each gene across all
networks. Additionally, we embed the positional encodings of the input gene networks to better identify the position of each unique
gene within the graph. These embeddings are integrated into graph transformer-based model, termed GT-GRN, for GRN inference.
The GT-GRN model effectively utilizes the topological structure of the ground truth network while incorporating the enriched encoded
information. Experimental results demonstrate that GT-GRN significantly outperforms existing GRN inference methods, achieving
superior accuracy and highlighting the robustness of our approach.
\end{abstract}


\section{Introduction}
The key aspect of Systems Biology seeks to understand the bigger picture of the complex biological systems that focuses to extract the relevant biological information of an organism at the cellular level. Biological components, such as genes, interact with each other to reconstruct gene regulatory networks (GRNs) from observational gene expression data \cite{bellot2015netbenchmark}. This process unveils a complex web of interactions, shedding light on the underlying patterns that govern gene regulation. Inference of GRNs from gene expression data is crucial for understanding the multifactorial molecular interaction patterns among genes. A gene network consists of interlinked genes, where the expression of each gene influences the activity of other genes in the network \cite{de2010gene}. An effective approach to describing gene regulatory networks (GRNs) involves the use of graphical and mathematical modeling, often grounded in graph-theoretic formalism to capture the complex interactions between genes. Formally, a GRN is represented as a network of nodes and edges, where the nodes represent genes, and edges represent the regulatory interactions between them \cite{guzzi2020biological}. GRN inference involves predicting or computationally guessing the connections among macromolecules by analyzing their relative expression patterns.

Technologies such as DNA Microarray \cite{shyamsundar2005dna}, Single-Cell RNA Sequencing (scRNA-seq) \cite{kolodziejczyk2015technology}, and Single-Nucleus RNA Sequencing (snRNA-seq) \cite{grindberg2013rna} have revolutionized transcriptomics by offering diverse and detailed insights into gene expression. Although each of these technologies has its unique strengths, they also come with certain limitations. Common limitations include noisy gene expression data, which often complicates the inference process. Furthermore, the dynamic and non-linear nature of gene-gene regulatory interactions present a significant challenge, as traditional or linear methods often fail to capture the complex relationships comprehensively. In addition, gene regulatory networks tend to be sparse, further reducing the overall accuracy of inference methods. In the case of single-cell technologies, significant dropout events introduce a large number of zero counts in the expression matrix, adding another layer of complexity \cite{talwar2018autoimpute}. These challenges highlight the need for more robust and sophisticated approaches to effectively analyze and interpret gene expression data.

A plethora of supervised and unsupervised GRN inference methods have been developed to uncover intricate relationships within gene networks \cite{huynh2019gene, jha2020prioritizing}. Early efforts focused on simpler techniques, such as correlation analysis, mutual information-based methods, differential equations, and more. Significant effort has been devoted to understand the intricate relationships within gene networks \cite{roy2014reconstruction, sebastian2023generic, sebastian2025network}. However, many methods come with inherent limitations. Therefore, a more reliable GRN inference technique may prove more effective in this case. Numerous intelligent computational techniques have been developed to address this issue.

Over the years, significant progress has been made in developing approaches that incorporate machine learning techniques to infer the network from gene expression \cite{dewey2013gene}. A recent trend in network-science has gained significant momentum in modeling graph-based applications powered by graph neural networks (GNNs). For example, Wang et al., propose GRGNN\cite{wang2020inductive} method to reconstruct GRNs from gene expression data in a supervised and semi-supervised framework. The problem is formulated as a graph classification problem for GRN inference on DREAM5 benchmarks. Q-GAT\cite{zhang2023quadratic} proposes quadratic neurons based on dual attention mechanism for GRN inference. The model is validated by introducing adversarial perturbations to the gene expression data on \textit{E. coli} and \textit{S. cerevisiae} datasets. Huang et al.,\cite{huang2023miggri} propose a GNN based model called MIGGRI for GRN inference using spatial expression images that capture gene regulation from multiple images. 

Another line of work also concentrates on Transformer-based architectures for GRN inference. TRENDY \cite{tian2024trendy} leverages transformer models to construct a pseudo-covariance matrix as part of the WENDY \cite{wang2024wendy} framework. Rather than generating gene regulatory networks (GRNs) from scratch, it enhances existing inferred GRNs. However, it does not incorporate additional structural side information into the inference process.  STGRNS \cite{xu2023stgrns} is an interpretable transformer-based method for inferring GRNs from scRNA-seq data. This method only considers two genes for gene regulation excluding the possibility of indirect regulation for prediction. GRN-Transformer \cite{shu2022boosting} utilizes multiple statistical features extracted from scRNA-seq data and uses inferred GRN extracted from a single inference algorithm PIDC\cite{chan2017gene}. 

Despite the advancements and superior performance of several methods, notable limitations persist. A primary concern is that many approaches rely solely on a single source of information, that is, gene expression data for inferring gene regulatory networks (GRNs), which is often insufficient for accurate and conclusive network prediction. Furthermore, some methods fail to utilize existing knowledge from previously inferred GRNs, limiting their potential to build upon existing insights. Moreover, several techniques overlook the integration of topological information, which is crucial for capturing the structural properties essential for robust GRN inference. 

Rather than focusing solely on each aforementioned issue, our approach adopts an integrated perspective—driven by the intuition that combining multiple complementary sources of information, beyond gene expression alone, can enhance the quality of GRN inference. We propose \emph{GT-GRN}, a novel approach that integrates the strengths of both unsupervised inference methods and supervised learning frameworks. Our method combines outcomes from the available inference techniques to minimize method-specific biases, ultimately deriving a more realistic and biologically meaningful GRN. To integrate multiple networks, rather than relying on GNN-based methods, which often suffer from over-smoothing when stacking multiple layers. We adopt a state-of-the-art NLP-based unsupervised approach that effectively captures and integrates information across networks.
\emph{GT-GRN} leverages the latest advancements in Graph Transformer models to enhance GRN inference. \emph{GT-GRN} integrates three distinct representations derived from  input expression networks: (1) topological features, which capture the structural properties of the network; (2) gene expression values, which are crucial for identifying gene interactions; and (3) the positional importance of genes, which reflects their functional relevance within the network. By fusing the multi-modal embeddings from diverse perspectives, our framework improves both the interpretability and predictive power of inferred GRNs, making it a robust solution for various biological applications.  \emph{GT-GRN} is adventageous in multiple sense. 

1) \emph{Multi-Networks Integration}: 
A key challenge in supervised Gene Regulatory Network (GRN) inference is the absence of ground-truth networks. True GRNs are often incomplete or unavailable, so we must rely on inferred networks as proxies. However, using a single inferred network can introduce bias or overlook critical interactions. We incorporate multiple networks inferred by different methods, harnessing their complementary strengths. While various inference models exist, each with its own set of advantages and limitations, combining these diverse sources allows us to leverage their shared strengths. This approach helps mitigate methodological bias, ultimately enhancing the confidence and accuracy of our GRN predictions.

2) \emph{Gene Expression Embedding}: Capturing meaningful representations of gene expression data through advanced embedding techniques can provide a richer understanding of the underlying regulatory mechanisms and improve GRN inference.  

3) \emph{Graph Transformer Frameworks}: Traditional GNNs rely on local message-passing mechanisms to infer graph structures. However, adopting graph transformer-based frameworks offers a more effective encoding strategy by leveraging global attention mechanisms, enabling better capture of complex regulatory relationships in GRNs.

The contributions of our present work are listed below: 

\begin{itemize}
    
    \item We capture the quantitative characteristics of gene expression profiles through an autoencoder that learns biologically meaningful latent representations, effectively summarizing complex gene activity patterns while preserving essential regulatory signals. ({\bf Section~\ref{sec:gene_embedding}}). 

    \vspace{1.5mm}
    
    \item We introduce a method to consolidate prior knowledge from multiple inferred GRNs by converting networks into text-like sequences, enabling a BERT-based masked language model to learn global gene embeddings that integrate structural information across all networks. ({\bf Section~\ref{sec:global_embedding}}). 

    \vspace{1.5mm}

    \item We propose a novel framework, \emph{GT-GRN}, which leverages attention mechanisms within a Graph Transformer model to learn rich gene embeddings by integrating multi-source data—including gene expression profiles, structured inferred knowledge, and graph positional encodings from the input graph. These unified gene embeddings effectively capture the underlying biological relationships between genes, facilitating enhanced Gene Regulatory Network (GRN) inference ({\bf Section~\ref{sec:gt}}).
    
    \vspace{1.5mm}

    \item We demonstrate that \emph{GT-GRN} effectively advances cell-type-specific gene regulatory network (GRN) reconstruction. Moreover, the superior quality of the learned embeddings enables their successful application to cell type annotation tasks, highlighting the model’s robustness and generalizability.

\end{itemize}

\section{Related Work}
With decades of effort within the research community dedicated to deciphering gene regulatory relationships from gene expression data, numerous methods have been proposed for reconstructing GRNs\cite{mochida2018statistical}. Traditional approaches include regression-based techniques \cite{haury2012tigress} and mutual information-based methods such as ARACNE \cite{margolin2006aracne}, MRNET \cite{meyer2007information}, and CLR \cite{faith2007large}, which assess statistical dependencies between genes to infer potential regulatory interactions. Supervised machine learning methods have also been explored for GRN inference. Support Vector Machines (SVMs) have been utilized to reconstruct biological networks through local modeling approaches \cite{bleakley2007supervised}. Extensions such as CompareSVM \cite{gillani2014comparesvm} and GRADIS \cite{razaghi2020supervised} further leverage classification-based frameworks to enhance network prediction accuracy.

With the advent of deep learning, more powerful and data-driven models have emerged. For instance, Daoudi et al. \cite{daoudi2019deep} proposed a deep neural network (DNN) model to infer GRNs from multifactorial experimental data. Turki et al. \cite{turki2016inferring} integrated both supervised and unsupervised learning techniques to perform link prediction on time-series gene expression data. Mao et al. \cite{mao2023gene} introduced a 3D convolutional neural network (CNN) model utilizing single-cell transcriptomic data, employing a novel labeling trick to enhance performance. GNE \cite{kc2019gne}, a graph-based deep learning framework, unified known gene interactions and expression profiles to robustly infer GRNs in a scalable manner. Other works, Teji et al.,\cite{teji2022predicting, teji2023graph} uses synthetic data using to evalaute various embedding models to evaluate for GRN inference in link prediction setup.

Recent approaches have increasingly explored the utility of graph neural networks (GNNs) to capture the structural and relational complexity of biological systems. DeepRIG \cite{wang2023inferring} emphasizes learning global regulatory structures by embedding entire graphs using a graph autoencoder, thereby capturing comprehensive latent representations. GMFGRN \cite{li2024gmfgrn} applies graph convolutional networks (GCNs) to factorize single-cell RNA-seq data into gene and cell embeddings, which are then used in a multilayer perceptron (MLP) for interaction prediction. GNNLink \cite{mao2023predicting} frames GRN inference as a link prediction problem by employing a GCN-based interaction graph encoder to capture and infer potential regulatory dependencies between genes.

Despite these advancements, many GRN inference methods continue to rely on a single source of data or even topological information, often emphasizing local or pairwise gene interactions. This narrow focus can limit the depth and breadth of biological insights. Although deep learning approaches—such as MLPs, CNNs, and GNNs—have enhanced inference performance, they frequently process each data modality in isolation, missing opportunities for deeper integration.

Present research is a step forward that proposes a novel graph transformer framework integrated with multiple sources of biological information for GRN inference. Unlike conventional methods that rely on convolution-based architectures, our approach utilizes graph-based attention mechanisms to effectively model complex regulatory relationships. A key strength of this framework lies in its ability to fuse diverse embeddings derived from gene expression data, input graph structures, and both existing and previously inferred regulatory networks. By jointly leveraging these complementary sources within a unified, the model enables more accurate and biologically meaningful inference of gene regulatory networks.

\section{Materials and Methodology}\label{sec:methodology}
This section outlines the methodology of the proposed \emph{GT-GRN} framework for gene regulatory network (GRN) inference, followed by a description of the datasets used for evaluation. The framework is composed of three key modules: (1) encoding gene expression profiles as embedding features using unsupervised deep learning; (2) extracting global gene embeddings through multi-network integration; and (3) capturing graph positional encodings from the input network structure. These complementary representations—gene expression embeddings, prior gene representations, and graph positional encodings—are fused to enhance GRN interaction prediction within a graph transformer model. The effectiveness of our approach is then evaluated using publicly available gene expression datasets.

 \subsection{Gene Expression Feature Encoding}\label{sec:gene_embedding}
\begin{figure*}
    \centering
    \includegraphics[width=15cm, height=6cm]{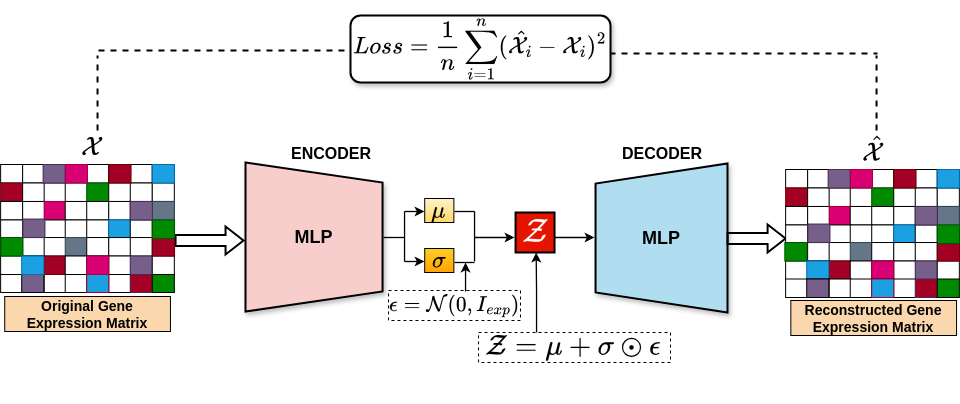}
    \caption{\textbf{Gene Expression Embeddings via Variational Autoencoder (VAE)}}
    \label{fig:feature_generation}
\end{figure*}

Understanding the dynamics of gene expression data involves unraveling the patterns and relationships embedded within large-scale datasets. As profiling technologies rapidly advance, gene expression data has become increasingly complex and high-dimensional. In such settings, traditional linear models often fall short in capturing the intricate patterns within the data. Unsupervised deep learning methods hold significant potential for uncovering meaningful biological signals and patterns from gene expression data. By leveraging such methods to non-linearly encode expression profiles, we can extract richer and more informative representations that capture the underlying biological complexity. Here, we employ  variational autoencoder (VAE) \cite{kingma2013auto} to encode gene expression data into compact, meaningful feature representations. Figure \ref{fig:feature_generation} illustrates the overview VAEs for encoding gene expressions. Variational Autoencoders (VAEs) are a probablistic deep generative class of neural networks designed to reconstruct input data by learning a compressed, low-dimensional representation that effectively characterizes the input. Using VAE for gene expression encoding, we can efficiently capture complex expression dynamics and generate compact feature representations for further analysis and downstream tasks. VAEs are a powerful framework for unsupervised learning and generally comprise two interconnected components: an \emph{encoder} and a \emph{decoder}. 


\begin{itemize}
    \item {\bf Encoder} : maps the input gene expression matrix $\mathcal{X}$ to a latent representation space $\mathcal{Z}$. It approximates the posterior distribution $\alpha(\mathcal{Z}|\mathcal{X})$ using a neural network. The encoder outputs the parameters, i.e., the mean and variance of a multivariate Gaussian distribution $\beta(\mathcal{Z}|\mathcal{X})$, that serves as an approximation of the true posterior $\alpha(\mathcal{Z}|\mathcal{X})$. This process captures the underlying biological variability and regulatory patterns among genes.

    \item {\bf Decoder} : takes a sample $\mathcal{Z}$ from the latent space and maps it back to the original gene expression space, generating a reconstructed matrix $\hat{\mathcal{X}}$. This is modeled by the likelihood $\alpha(\mathcal{X}|\mathcal{Z})$, which represents the probability of generating the observed gene expression profiles $\mathcal{X}$ given the latent variables $\mathcal{Z}$. The decoder, implemented via a neural network, learns to reconstruct biologically plausible gene expression patterns from the learned latent representations.
\end{itemize}

During training, the VAE aims to learn the parameters of the encoder and the decoder network parameters by maximizing the Evidence Lower Bound (ELBO) which is given by : 

\begin{equation}
    \text{ELBO} = \mathbb{E}_{\beta(\mathcal{Z}|\mathcal{X})} [\log \alpha(\mathcal{X}|\mathcal{Z})] - \text{KL}[\beta(\mathcal{Z}|\mathcal{X}) || \alpha(\mathcal{Z})]
\end{equation} 

where, $\mathbb{E}_{\beta(\mathcal{Z}|\mathcal{X})} [\log \alpha(\mathcal{X}|\mathcal{Z})]$ is the reconstruction term that reconstructs the input gene expression data $\mathcal{X}$ given the latent representation $\mathcal{Z}$.   $\text{KL}[\beta(\mathcal{Z}|\mathcal{X}) || \alpha(\mathcal{Z})]$ is the KL (Kullback-Leibler) divergence that quantifies the distance between the approximate posterior $\beta(\mathcal{Z}|\mathcal{X})$ and the prior distribution $\alpha(\mathcal{Z})$.   

Since sampling from learned distributions is inherently non-differentiable, it hinders the use of gradient-based optimization during backpropagation. To address this, the reparameterization trick introduces a differentiable transformation by expressing the random variable as 
$\mathcal{Z} = \tau(\mathcal{X}, \epsilon)$, where, $\tau$ is a deterministic function and $\epsilon$ is an auxiliary noise variable drawn from a fixed, independent distribution. The above problem can be re-written as: 

 $$  \mathcal{Z} \sim  \beta(\mathcal{Z}|\mathcal{X}^{(i)}) = \mathcal{N}(\mathcal{Z} ; \mu^{(i)}, \sigma^{2(i)} I_{exp})$$

$$\mathcal{Z} = \mu +  \sigma \odot  \epsilon \text{; reparamterization trick}$$ 

Where, $\epsilon = \mathcal{N}(0, I_{exp})$, $\odot$ is the element-wise product and $I_{exp}$ is the identity matrix, which serves as the covariance matrix. 


\subsection{Global Embeddings via Multi-Network Integration of the inferred GRNs}\label{sec:global_embedding}
\begin{figure*}
    \centering
    \fbox{\includegraphics[width=0.98\linewidth]{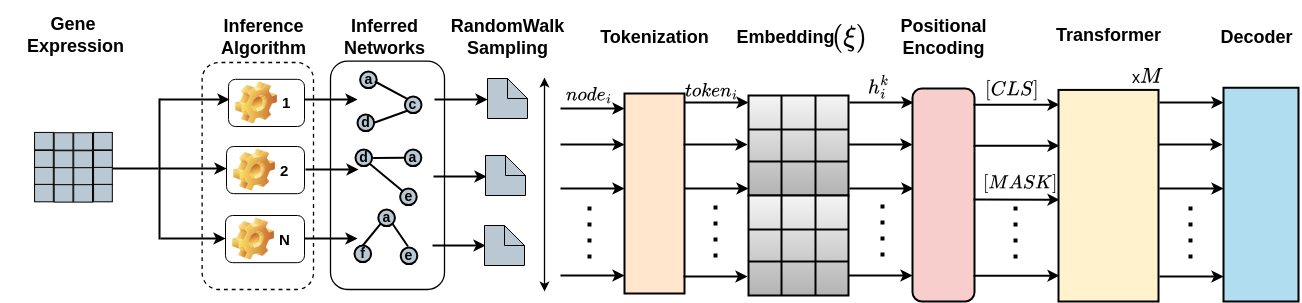}}
    \caption{\textbf{Global Gene Embeddings via Multi-Network Integration}. \emph{Gene expression data is processed through inference algorithms to generate networks. Every network is  sampled via random walks to produce node sequences. Each sequence starts with a special [CLS] token. These sequences are tokenized and embedded, then passed to a transformer model resulting in global embedding of each node or gene considering the context of all input networks in a joint learning setting. The model is trained by masking nodes in sequences and predicting them. Final node/gene embeddings are extracted from the embedding layer.}}
    \label{fig:global_emb}
\end{figure*}

We integrate multi-network information to understand gene interactions as prior knowledge. Integrating data from diverse inference methods, each with unique strengths and limitations, provides a holistic and reliable view of the network. This approach overcomes the shortcomings of relying on a single method, enabling robust downstream analysis. Figure \ref{fig:global_emb} illustrates the workflow. 

\subsubsection{Unsupervised Network Integration via Random Walks and Transformers}

We present an unsupervised learning approach for integrating multiple networks to generate global embeddings. Let a graph \(\mathcal{G} = (\mathcal{V}, \mathcal{E})\) represent an inferred network, where \(\mathcal{V}\) denotes the set of nodes and \(\mathcal{E}\) represents the set of edges. The graph is characterized by its adjacency matrix \(\mathcal{A} \in \mathbb{R}^{|\mathcal{V}| \times |\mathcal{V}|}\), where each entry \(\mathcal{A}_{ij}\) reflects the relationship between nodes \(v_i\) and \(v_j\). Specifically, \(\mathcal{A}_{ij} = 1\) if an edge \(e_{v_i, v_j} \in \mathcal{E}\) connects \(v_i\) and \(v_j\), and \(\mathcal{A}_{ij} = 0\) otherwise, indicating no direct connection. 


We consider a collection of \(C\) networks, \(\{G_1, G_2, \dots, G_C\}\), sharing the same set of \(n\) nodes but differing in the number of edges in each network. We capture the structural information of the networks by converting them into text-like sequences using random walks, similar to node2vec \cite{grover2016node2vec}. The walks are  encoded through an embedding matrix \(\xi \in \mathbb{R}^{n \times d_n}\), where \(n\) is the size of the vocabulary (total nodes across all networks) and \(p\) is the desired embedding dimension. Positional encodings are used to account for node order as described in \cite{vaswani2017attention}:

\begin{equation}
PE_{\text{seq}}(\text{pos}, i) = 
\begin{cases} 
\sin\left(\dfrac{\text{pos}}{10000^{i/p}}\right), & \text{if } i \bmod 2 = 0, \\[1ex]
\cos\left(\dfrac{\text{pos}}{10000^{(i-1)/p}}\right), & \text{otherwise}.
\end{cases}
\end{equation}

Here, \(PE_{seq}{\text{(pos, i)}}\) represents the \(i\)-th coordinate of the position encoding at sequence position \(pos\). These encodings are concatenated with the original input features or the embedding matrix.

The embedding matrix and the decoding layer are initialized with uniform random values, while the transformer layer is initialized using Xavier's initialization \cite{devlin2018bert}. During training, all parameters are updated. Each sequence begins with a special classification token \([CLS]\), while other tokens correspond to node-specific vectors from the embedding matrix. The final hidden state of the \([CLS]\) token for a given sequence serves as the sequence representation.

\subsubsection{Masked Language Learning with BERT}
We utilize the Masked Language Modeling (MLM) approach as implemented in BERT \cite{devlin2018bert}. At its core, this method employs a transformer encoder composed of \( M\) identical blocks. Each block includes a self-attention mechanism followed by a feedforward neural network, as described in \cite{vaswani2017attention}.  

Let \( F = [f_1, f_2, f_3, \dots, f_n] \) denote an input sequence of \( n \) tokens, where each token is represented by a \( p \)-dimensional vector. A self-attention layer processes this sequence using the following transformation:  

\begin{equation}
\text{Attention}_{\text{seq}}(Q_{\text{seq}}, K_{\text{seq}}, V_{\text{seq}}) = 
\text{softmax}\left(\frac{Q_{\text{seq}}K_{\text{seq}}^\top}{\sqrt{d_{n_{k_{\text{seq}}}}}}\right)V_{\text{seq}},
\end{equation}

where, \( Q_{\text{seq}} = W_{q\_\text{seq}}F \), \( K_{\text{seq}} = W_{k\_\text{seq}}F \), and \( V_{\text{seq}} = W_{v\_\text{seq}}F \). Here, \( W_{q\_\text{seq}} \), \( W_{k\_\text{seq}} \), and \( W_{v\_\text{seq}} \) are learnable matrices that project the input into query, key, and value spaces of node sequences, respectively. \( d_{n_{k_{\text{seq}}}} \) is the dimension of the key vectors.

The feedforward layer, applied independently to each token, performs the transformation:  

\begin{equation}
FFN(\xi) = \text{ReLU}(\xi W_{seq1} + b_{seq1})W_{seq2} + b_{seq2},
\end{equation}

where \( W_{seq1} \), \( W_{seq2} \) are learnable matrices, and \( b_{seq1} \), \( b_{seq2} \) are bias vectors. $FFN$ is the feed-forward network and $\xi$ is the global gene embeddings.  

The MLM task involves masking a random subset of input tokens and predicting their identities based on the remaining context. This encourages the model to capture bidirectional contextual relationships within sequences. Specifically, we mask \( 20\% \) of the tokens (representing nodes) in each sequence and train the model to recover the masked tokens using a cross-entropy loss function:  

\begin{equation}
L_{bert} = -\sum_{b=1}^{B} \sum_{ln=1}^{Ln} \sum_{cl=1}^{Mc} 1_{\{b, ln \in \text{mask}\}} \cdot y_{b,ln,cl} \log p_{b,ln,cl},
\end{equation}

where, \( B \) is the batch size, \( Ln \) is the sequence length and \( Mc \) is the number of classes (total number of possible tokens that can be predicted).  $y_{b,ln,cl}$ is a binary indicator equal to \( 1 \) if the correct class of token \( ln \) in batch \( b \) is \( cl \), and \( p_{b,ln,cl} \) is the predicted probability for this classification. The final embeddings are extracted from the embedding layer represented as $\xi$. 

This enables the model to learn rich contextual embeddings for nodes, capturing both structural and positional relationships within the networks.

\subsection{Graph Transformer for GRN inference}\label{sec:gt}

\begin{figure}
    \centering
    \fbox{\includegraphics[width=8.66cm, height=19cm]{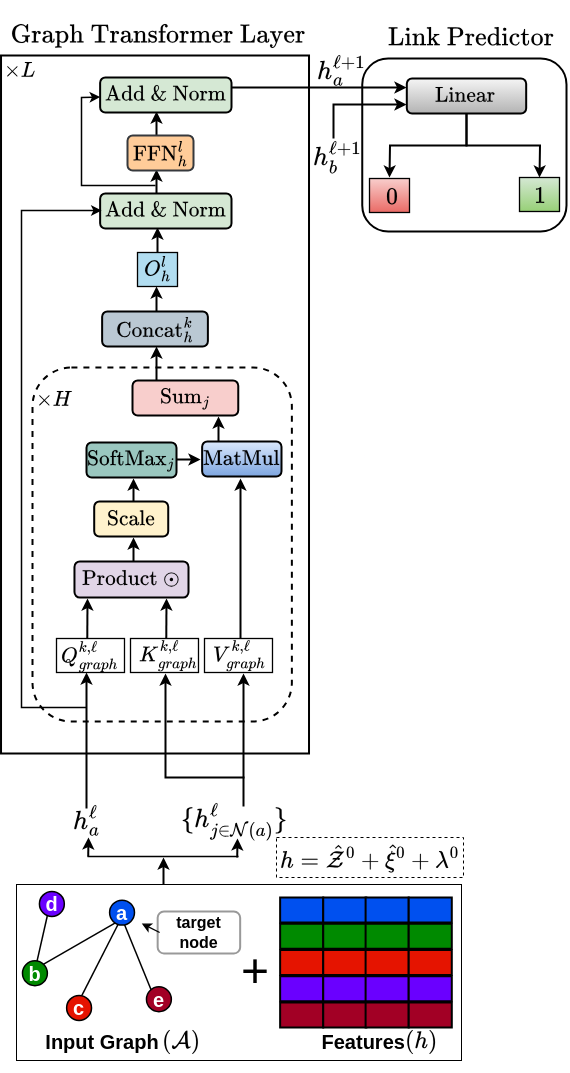}}  
    \caption{\textbf{Architecture diagram of \emph{GT-GRN}.} \emph{Graph Transformer layer operates on the input graph $\mathcal{A}$ with its corresponding features $h$. The input features consists of gene expression embeddings , graph positional encodings and the global gene embeddings. It operates to compute the node embeddings for a particular node $\textbf{a}$ after passing through multiple layers $L$ to produce representations at the next layer $h_a^{\ell+1}$. Link prediction is done via a separate link predictor module that takes two node embeddings say $h_a^{\ell+1}$ and $h_b^{\ell+1}$ to predict a link between them.}}
    \label{fig:GT_GRN}
\end{figure}

After deriving features from available GRNs and gene expression data, we utilize the Graph Transformer (GT) to learn comprehensive representations by injecting the underlying regulatory structure. Since GT is specifically designed to model complex dependencies in graph-structured data, it effectively captures gene-gene interactions based on attention mechanisms, making it well suited for GRN-based representation learning. \emph{GT-GRN} is illustrated in Figure \ref{fig:GT_GRN}. 

\subsubsection{Graph Positional Encodings}
NLP-based Transformers are supplied with Positional Encodings. At the heart of GT, graph positional encodings hold a special place which is important for encoding node positions. From the available graph structure, we make use of Laplacian eigenvectors and use them as graph positional encoding ($PE_{graph}$) information. This is helpful to encode distance-aware information, i.e., nearby nodes would have similar positional features and vice versa. Eigenvectors are defined as the factorization of the graph Laplacian matrix:

\begin{equation}
\Delta = I_{graph} - D^{-\frac{1}{2}} \mathcal{A} \thinspace D^{-\frac{1}{2}} = U \Lambda U^{T}
\end{equation}

where $\mathcal{A}$ is the input adjacency matrix of $n \times n$ dimensions, $I_{graph}$ is the identity matrix of size $n \times n$. $D$ is the degree matrix, $\Lambda$ is the eigenvalues and $U$ are the eigenvectors. We then use the $p$ smallest significant eigen-vectors of a node as its positional encoding, which is denoted by $\lambda_i$ for node $i$.

\subsubsection{Input to \textit{GTGRN}}
The input to the Graph Transformer (GT) layer is the graph structure \( \mathcal{G} \) and its associated features \( h \). The features \( h \) are constructed as a combination of gene expression embeddings (\( \mathcal{Z} \)), global gene embeddings (\( \xi \)), and graph positional embeddings (\( \lambda \)), which are derived from the graph \( \mathcal{G} \). 

\begin{itemize}
    
\item  For gene expression embeddings, each gene \( \mathcal{Z}_i \in \mathbb{R}^{d_n \times 1} \) is passed through a linear projection layer to embed it into a \( d \)-dimensional space. 

\begin{equation}
    \hat{\mathcal{Z}}^0_i = S^0 \mathcal{Z}_i + s^0 
\end{equation}

where, \( S^0 \in \mathbb{R}^{d \times d_n} \) is a learnable weight matrix, \( s^0 \in \mathbb{R}^d \)is the learnable bias vector, and \( \hat{\mathcal{Z}}^0_i \in \mathbb{R}^d \) is the projected embedding.

\vspace{0.2cm}

\item For the global gene embeddings, each vector \( \xi_i \in \mathbb{R}^{d_n \times 1} \) is also embedded into a 
$d$ dimensional
 space using a separate linear projection layer. The transformation is defined as:
\begin{equation}
\hat{\xi}^0_i = T^0 \xi_i + t^0\end{equation}

where, \( T^0 \in \mathbb{R}^{d \times d_n} \) is a learnable weight matrix, \( t^0 \in \mathbb{R}^d \) is the learnable bias vector. 

\vspace{0.2cm}

\item The graph positional encodings are extracted from the input graph $\mathcal{G}$. For a particular gene's positional encoding  \( \lambda_i \in \mathbb{R}^{d_n} \) is embedded into \( d \)-dimensional space using a linear projection layer which is given by:

\begin{equation}
    \lambda^0_i = U^0 \lambda_i + u^0    
\end{equation}

where, \( U^0 \in \mathbb{R}^{d \times d_n} \) is a learnable weight matrix, \( u^0 \in \mathbb{R}^d \) is a learnable bias vector. 

\end{itemize}


Finally, the gene expression embeddings \( \hat{\mathcal{Z}}^0 \in \mathbb{R}^{n \times d} \), global gene embeddings \( \hat{\xi}^0 \in \mathbb{R}^{n \times d} \), and graph positional embeddings \( \lambda^0 \in \mathbb{R}^{n \times d} \) are each projected into a shared \( d \)-dimensional space through separate linear transformation layers. These projected representations are then summed element-wise (often called fusion by summation) to form the final node features \( h \in \mathbb{R}^{n \times d} \):

\begin{equation}
    h = \hat{\mathcal{Z}}^0 + \hat{\xi}^0 + \lambda^0
\end{equation}

where, $h$ is the unified, per‑gene embedding that fuses its expression profile, its topological position in the regulatory network, and a dataset‑wide global context.
This final representation \( h \) is then injected into the GT layer along with the adjacency matrix \( \mathcal{A} \) of the input graph \( \mathcal{G} \).

\subsubsection{Graph Transformer Layer} The node update in the GT at layer ${\ell}$ is defined as follows: 

\begin{equation}
\hat{h}_{i}^{\ell+1} = O_h^{\ell} \, \bigg\|_{k=1}^{H} \left( \sum_{j \in \mathcal{N}(i)} w_{ij}^{k,\ell} V_{graph}^{k,\ell} h_{j}^{\ell} \right)
\end{equation}

\begin{equation}
\text{where,} \thickspace  w_{ij}^{k,\ell} = \text{softmax}_j \left( \frac{Q_{graph}^{k,\ell}h_{i}^{\ell} \cdot K_{graph}^{k,\ell} h_{j}^{\ell}}{\sqrt{d_k}} \right),
\end{equation}

and $d_k = d/H$, $k=1 \thickspace \text{to} \thickspace H$ denotes number of attention heads, and $\bigg\|$ denotes the contatenation of the number of heads.. $Q_{graph}^{k,{\ell}}$, $K_{graph}^{k,{\ell}}$, $V_{graph}^{k,{\ell}} \in \mathbb{R}^{d_k \times d}$, $O_h^{\ell} \in \mathbb{R}^{d \times d}
$,  The attention outputs $\hat{h}_i^{{\ell}+1}$ are then passed to a Feed Forward Network (FFN) preceded and succeeded by residual connections and normalization layers as:

\begin{equation}
    \hat{\hat{h}}_{i}^{\ell+1} = \text{Norm} \left( h_{i}^{\ell} + \hat{h}_{i}^{\ell+1} \right)
\end{equation}

\begin{equation}
    \hat{\hat{\hat{h}}}_{i}^{\ell+1} = W_{graph2}^{\ell} \, \text{ReLU} \left( W_{graph1}^{\ell} \hat{\hat{h}}_{i}^{\ell+1} \right)
\end{equation}


\begin{equation}
    h_{i}^{\ell+1} = \text{Norm} \left( \hat{\hat{h}}_{i}^{\ell} + \hat{\hat{\hat{h}}}_{i}^{\ell+1} \right)
\end{equation}

where $W_{graph1}^{\ell} \in \mathbb{R}^{2d \times d}$, $W_{graph2}^{\ell} \in \mathbb{R}^{d \times 2d}$, and $\hat{\hat{h}}_i^{{\ell}+ 1}$, $\hat{\hat{\hat{h}}}_i^{\ell + 1}$ are the intermediate representations. Norm could be either Layer-Norm\cite{ba2016layer} or BatchNorm\cite{ioffe2015batch}.

\subsubsection{Link Prediction with learned representations}
The final module is designed to predict edges between nodes using the learned node representations \( h^{\ell+1} \) obtained from the GT layer. The module takes as input the embeddings \( h^{\ell+1} \in \mathbb{R}^{n \times d} \), where \( n \) is the number of nodes and \( d \) is the embedding dimension, along with an edge index representing node pairs. For each edge \( (i, j) \), the embeddings of the source node \( h^{\ell+1}_i \) and the destination node \( h^{\ell+1}_j \) are extracted and concatenated to form a feature vector. This vector is passed through a decoder network consisting of a multi-layer perceptron (MLP) with a hidden layer, ReLU activation, and an output layer, which reduces the concatenated vector to a scalar. The scalar represents the predicted likelihood of an edge between the nodes \( i \) and \( j \). By utilizing the updated node embeddings \( h^{\ell+1} \), this module effectively learns to identify and score potential edges in the graph.

Next, we discuss the experimental setup used to demonstrate the superiority of \textit{GT-GRN.}

\subsection{Experimental Setup}
The performance of \emph{GT-GRN} is evaluated on Linux based NVIDIA RTX A3000 GPU as the computing machine. The deep learning libraries used here are pytorch\footnote{\url{https://pytorch.org/}}, dgl \footnote{\url{https://www.dgl.ai/}}, scikit-learn\footnote{\url{https://scikit-learn.org/stable/}}, Pytorch Geometric \footnote{\url{https://pytorch-geometric.readthedocs.io/en/latest/index.html\#}}. 
\subsection{Datasets}
We establish our findings on two single-cell RNA-sequencing (scRNA-seq) human cell types: human embryonic stem cells (hESC) \cite{yuan2019deep} and mouse embryonic stem cells (mESC) \cite{camp2017multilineage} from BEELINE \cite{pratapa2020benchmarking} study. The cell-type-specific ChIP-seq ground-truth networks are used as a reference for these datasets. Additionally, we use the synthetic expression profiles generated using GeneNetWeaver (GNW) \cite{schaffter2011genenetweaver}, a simulation tool developed for DREAM\footnote{Dialogue on Reverse Engineering Assessment and Methods} along with their corresponding ground-truth networks. The details of the datasets have been discussed in Table \ref{tab:dataset}.

\begin{table}[h]
\centering
\caption{\textbf{Dataset statistics}}\label{tab:dataset}
\begin{adjustbox}{height = 0.44in, width=3.2in}

\begin{tabular}{|c|c|c|c|c|c|c|c|}
\hline
\rowcolor[HTML]{D6DBDF}
\textbf{Species/Cell Types} & \textbf{Type} &  \textbf{Source}      & \textbf{$\mathcal{V}$}  &  \textbf{$\mathcal{E}$}    \\
\hline
Yeast   & Microarray & GNW        &   4000   &  11323   \\ 
hESC-500 & scRNA-seq & BEELINE   & 910 & 3940  \\
mESC-500 & scRNA-seq & BEELINE  & 1120  & 20923   \\
hESC-1000 & scRNA-seq & BEELINE  &  1410 & 6139   \\
mESC-1000 & scRNA-seq & BEELINE   & 1620  &  30254  \\

\hline
\end{tabular}
\end{adjustbox}
\end{table}

\subsubsection{Preprocessing of raw data}
We preprocess the raw scRNA-seq data using an established method \cite{pratapa2020benchmarking} to handle redundancy. We filter out low-expressed genes and prioritized the variable ones. Primarily, the genes expressed in less than 10\% of cells were removed. Then, we computed the variance and P-values for each gene, selecting those with P-values below 0.01 after Bonferroni correction. Gene expression levels were log-transformed for normalization. This yielded a feature matrix \( \mathcal{X} \in \mathbb{R}^{n \times m} \), where \( n \) is the number of genes and \( m \) is the number of cells. Furthermore, we adopt the approach of Pratapa et al.\cite{pratapa2020benchmarking} to assess performance across different network sizes. Specifically, we rank genes by variance and select the most variable transcription factors (TFs), along with the top 500 and 1000 genes with the highest variability.

\subsection{Baseline methods}
We evaluate the efficacy of \emph{GT-GRN} against the existing baselines methods commonly used for inferring GRNs are shown in Table \ref{tab:grn_methods}.

\begin{table*}[h]
\centering
\caption{\bf{Summary of Gene Regulatory Network (GRN) Inference Methods classified by Category.}}

\begin{adjustbox}{max width=\textwidth}
\renewcommand{\arraystretch}{1.3}
\begin{tabular}{|l|l|p{4.3cm}|l|}
    \hline
     \rowcolor[HTML]{D6DBDF} \textbf{Category} & \textbf{Method} & \textbf{Description} & \textbf{Source} \\
    \hline
    
    \multirow{3}{*}{\textbf{Graph Neural Network}} 
    & \textbf{GNNLink} \cite{mao2023predicting} &
    \begin{tabular}[c]{@{}c@{}}
    Uses a GCN-based interaction \\ graph encoder to capture \\ gene expression patterns. \end{tabular} & \url{https://github.com/sdesignates/GNNLink} \\
    \cline{2-4}
    
    & \textbf{GENELink} \cite{chen2022graph} & 
    \begin{tabular}[c]{@{}c@{}}
    Leverages a graph attention \\ network (GAT)  to infer GRNs \\ via attention mechanisms. \end{tabular} & \url{https://github.com/zpliulab/GENELink}
     \\
     \cline{2-4}
    & \textbf{GNE} \cite{kc2019gne} & 
    \begin{tabular}[c]{@{}c@{}}
    Uses an multi-layer perceptron \\ (MLP) to encode gene  expression \\ profiles and network  topology \\ for predicting gene regulatory links. \end{tabular} & \url{https://github.com/kckishan/GNE} \\
    \hline
    \multirow{4}{*}{\textbf{Mutual Information}} 
    & \textbf{ARACNE} \cite{margolin2006aracne} & 
    \begin{tabular}[c]{@{}c@{}}
    Infers networks based on \\ Adaptive Partitioning (AP) and \\ Mutual  Information  (MI) approach. \end{tabular} & \url{https://bioconductor.org/packages/release/bioc/html/minet.html} \\
    \cline{2-4}
    & \textbf{BC3NET} \cite{de2012bagging} &
    \begin{tabular}[c]{@{}c@{}}
    An ensemble technique derived \\ from the C3NET algorithm that \\  employs a bagging \\ approach for inferring GRNs. \end{tabular} & \url{https://cran.r-project.org/web/packages/bc3net/index.html} \\
    \cline{2-4}
    & \textbf{C3MTC} \cite{de2011influence} & 
    \begin{tabular}[c]{@{}c@{}}
    Infers gene interaction networks \\ where edge weights are \\ defined by corresponding \\ mutual information values. \end{tabular} & \url{https://cran.r-project.org/web/packages/bc3net/index.html} \\
    \cline{2-4}
    & \textbf{C3NET} \cite{altay2010inferring} & 
    \begin{tabular}[c]{@{}c@{}}
    Leverages mutual information \\ estimates and a maximization \\ step  to  efficiently capture \\ causal  structural information \\ in the data. \end{tabular} & \url{https://cran.r-project.org/web/packages/c3net/index.html} \\
    \hline
    \textbf{Feature Selection} 
        & \textbf{MRNET} \cite{meyer2007information} & 
        \begin{tabular}[c]{@{}c@{}}
        Applies a series of supervised \\ gene selection procedures  \\ using the MRMR  \\ (maximum relevance/ \\minimum redundancy) principle.\end{tabular} & \url{https://bioconductor.org/packages/release/bioc/html/minet.html} \\
    \hline
    \multirow{2}{*}{\textbf{Ensemble Tree-Based}} 
    & \textbf{GRNBOOST2} \cite{moerman2019grnboost2} &
    \begin{tabular}[c]{@{}c@{}}
    A fast GRN inference algorithm \\  using Stochastic Gradient Boosting\\  Machine (SGBM) regression. \end{tabular} & \url{https://github.com/aertslab/arboreto} \\
    \cline{2-4}
    & \textbf{GENIE3} \cite{huynh2010inferring} & 
    \begin{tabular}[c]{@{}c@{}}
    A classic GRN inference \\ algorithm using Random \\ Forest (RF) or \\ ExtraTrees (ET) regression.
    \end{tabular}
    & \url{https://github.com/aertslab/arboreto}\\
    \hline
\end{tabular}
\end{adjustbox}
\label{tab:grn_methods}
\end{table*}

\section{Results \& Analysis}
We report results using both single-cell and microarray gene expression datasets, selecting representative methods from each major category of gene regulatory network (GRN) inference techniques against \emph{GT-GRN}. These include mutual information-based methods, feature selection approaches, ensemble tree-based models, and graph neural network frameworks. This diverse selection allows us to comprehensively evaluate performance across different inference paradigms, ensuring a balanced comparison that highlights the strengths and limitations of each method.

\subsection{GRN Inference via Full Network Reconstruction}

\begin{figure*}[!ht]
\centering

\subfloat[]{\fbox{\includegraphics[width=8.4cm, height=6cm]{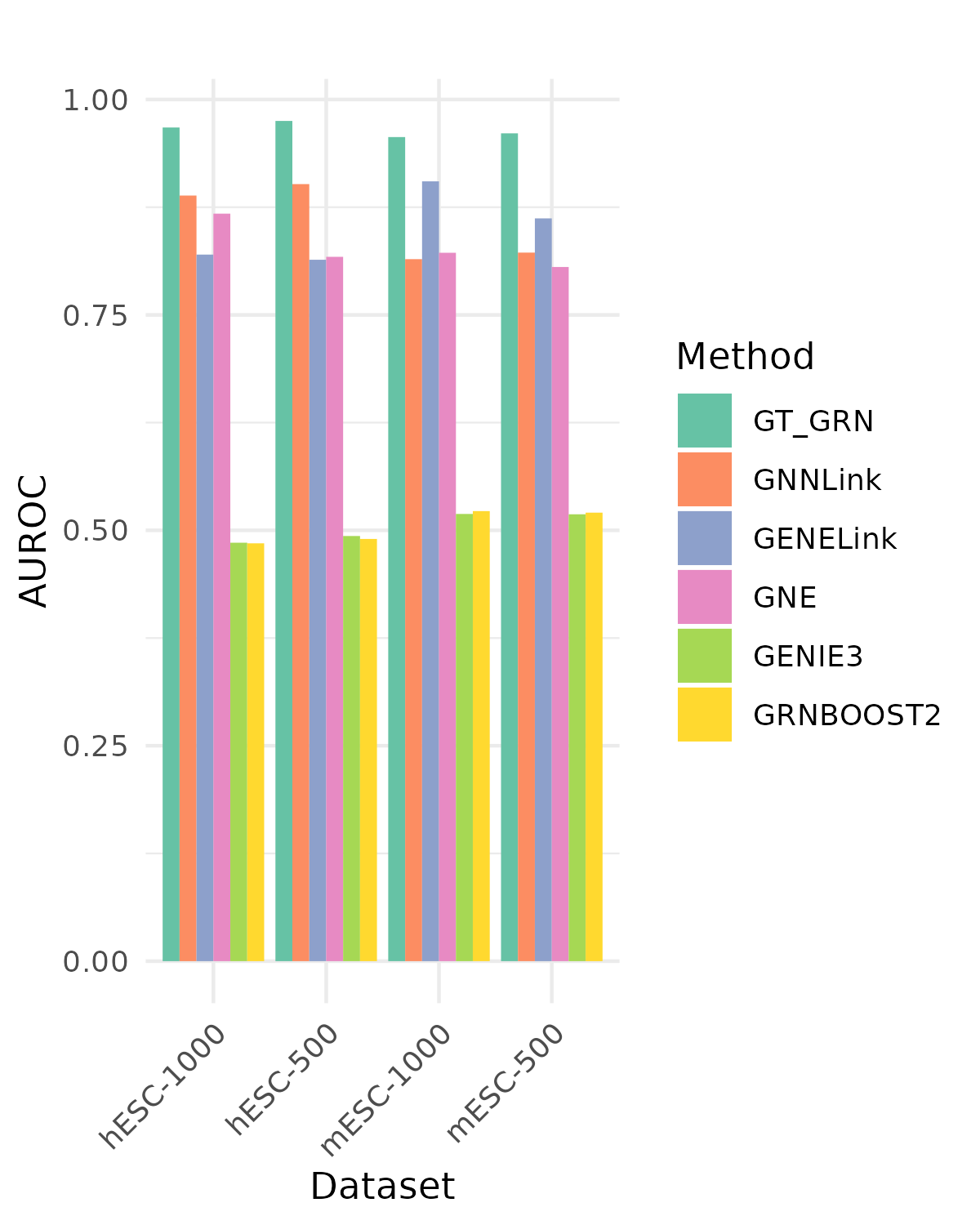}}}\label{fig:scRNA-seq_bar} \hspace{6mm}\subfloat[]{\fbox{\includegraphics[width=8.4cm,  height=6cm]{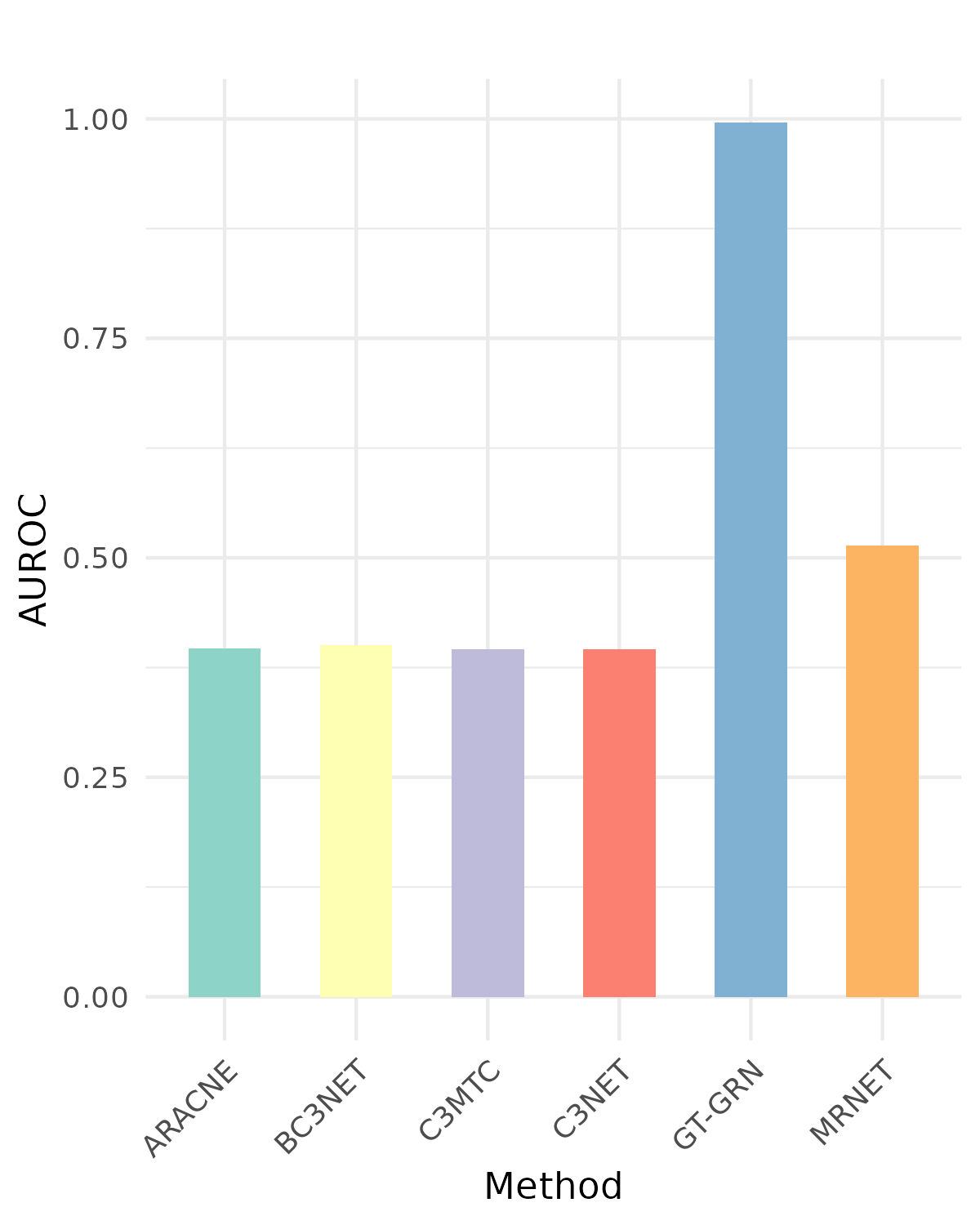}}}\label{fig:Yeast_bar}

\vspace{0.8cm}

\caption{\centering{\bf{Full network reconstruction performance of various methods for different datasets in terms AUROC score.}} \emph{(a) BEELINE’s scRNA-seq datasets and (b) GNW’s Yeast dataset. }}
\label{fig:network_reconstruction}

\end{figure*}

Fundamentally, Gene Regulatory Network (GRN) inference aims to reconstruct the entire regulatory network, capturing the full complexity of gene interactions. By striving for complete network reconstruction, it seeks to reveal the intricate web of regulatory relationships among all involved genes, reflecting the true, comprehensive regulatory architecture \cite{teji2024application}. We evaluate the effectiveness of \emph{GT-GRN} alongside existing methods designed for GRN inference. Following the similar motivation, we report these results for \emph{GT-GRN} and its baseline methods in Figure \ref{fig:network_reconstruction}. From the figure, the results of the full network reconstruction, highlight the comparative performance of different methods. For the scRNA dataset, the performance of \emph{GT-GRN} remains consistently higher with minor variations for all datasets. This suggests that the \emph{GT-GRN} method is robust for different cell types and sequencing depths. 

For the \textit{Yeast} (microarray) dataset, the \emph{GT-GRN} method significantly outperforms all other network inference methods, indicating its superior ability to capture gene interactions. Baseline methods such as ARACNE, BC3NET, C3MTC, C3NET, and MRNET perform at similar levels. In general, \emph{GT-GRN} appears to be the most effective method for the microarray dataset, while for scRNA expression profiles the method demonstrate stable performance across conditions in terms of AUROC score.

\subsection{GRN Inference via Link Prediction}


There exist many benchmark methods that treat GRN inference as a link prediction problem, focusing on identifying only a limited subset of interactions. We report results using this conventional approach. Table \ref{tab:auroc_auprc} reports the performance results for the candidate datasets in terms of \emph{Area
Under the Receiver-Operating Characteristic Curve} (AUROC) and \emph{Area Under the Precision-Recall Curve} (AUPRC) metrics. To assess the predictive performance of \emph{GT-GRN}, we present a comparison with various baseline methods. From the table it is clearly evident that the \emph{GT-GRN} consistently outperforms other methods in all datasets, achieving the highest AUROC and excelling in AUPRC, particularly for the mESC-1000 and mESC-500 datasets. GNE emerges as a strong contender, especially in the hESC-1000 dataset, where it achieves the highest AUPRC $0.9042$ and the AUROC of $0.9025$, indicating its effectiveness in balancing precision and recall. GENELink maintains strong performance across most datasets, while GNNLink performs well in AUROC but lags in AUPRC. In contrast, GENIE3 and GRNBOOST2 show consistently lower scores, indicating challenges in handling these complex datasets. Notably, GNE outperforms all methods in both AUROC and AUPRC for the hESC-1000 dataset, highlighting its scalability. For the Yeast-4000 dataset, \emph{GT-GRN} demonstrates superior performance compared to its baseline counterparts. Overall, \emph{GT-GRN} proves to be the most reliable across datasets, while GNE stands out for its precision, making them the top choices for biological network predictions in this study.

\subsection{Impact of Hyperparameters}
\begin{figure}[!h]
    \centering
\includegraphics[width=9.05cm, height=8.5cm]{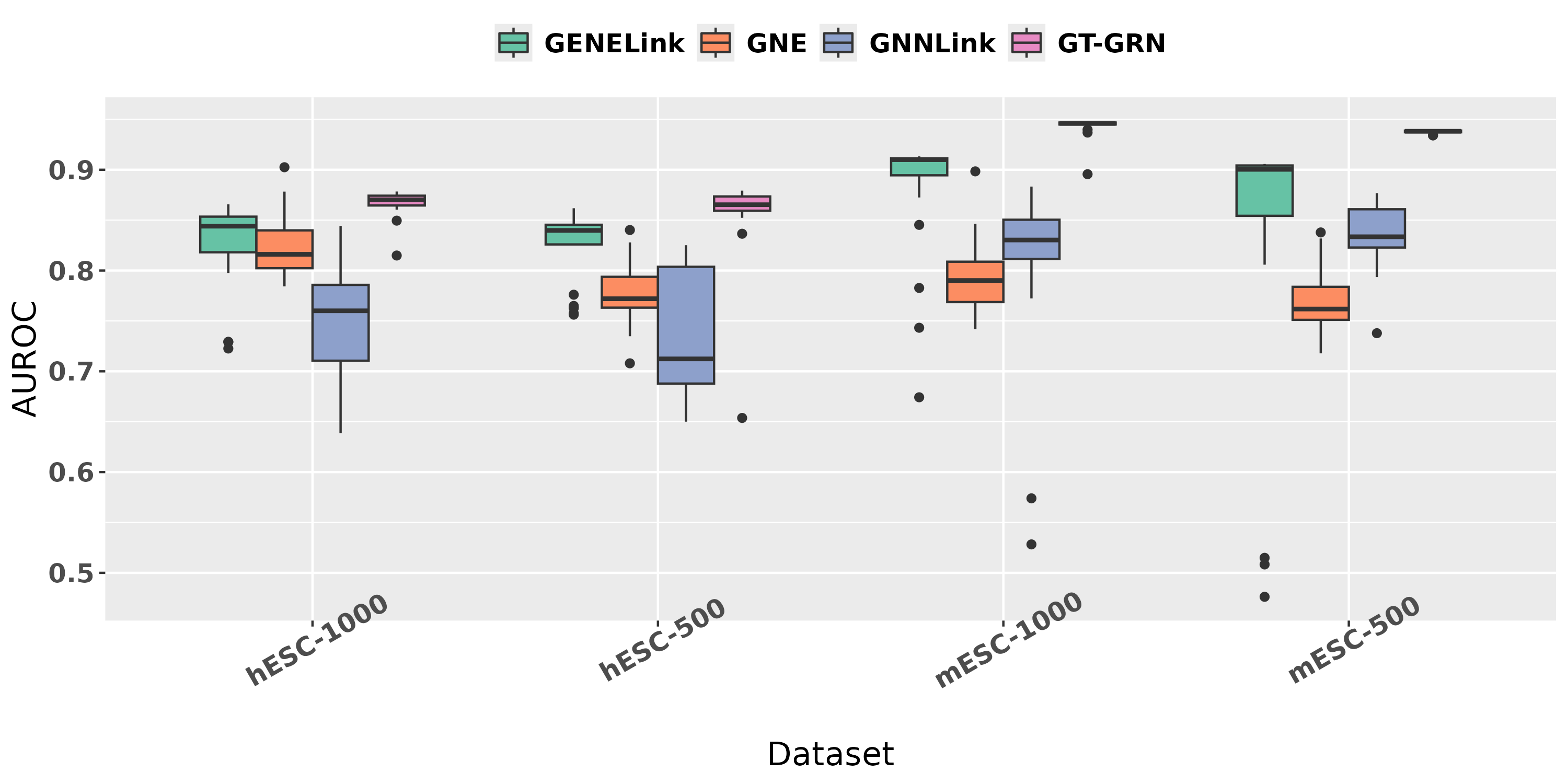}
    \caption{\bf{Overall hyper-parameter tuning plot for various models.}}
    \label{fig:HP_boxplot}
\end{figure}

Additionally, we aim to investigate the role of various hyperparameters (HPs) that influence the overall performance of the models. Optimal selection of HPs is a difficult task and time-consuming activity. We analyze the behavior of the learning models compared to \emph{GT-GRN} by selecting key hyperparameters from each baseline model to assess their impact on the overall performance of each model. Table \ref{tab: hyperparameters} describes the parameteric configurations that we tuned for each learning model with their respective explanations.  We summarize the overall impact of hyperparameter tuning for different models using a boxplot, as shown in Figure~\ref{fig:HP_boxplot}. This visualization presents a comprehensive comparison of our model, \emph{GT-GRN}, against several baseline approaches in terms of AUROC performance across multiple datasets.

A key observation is that \emph{GT-GRN} consistently achieves higher median AUROC scores with notably lower variance across datasets, highlighting its robustness and reliability under varying experimental conditions. Each model was tuned with its respective optimal hyperparameters, yet \emph{GT-GRN} exhibits both stability and effectiveness across settings, clearly outperforming the existing baselines on all candidate datasets.  In contrast, GNNLink shows a higher variance, particularly in the hESC-500 dataset, where its performance fluctuates significantly. GENELink and GNE display relatively stable performances, though both fall short of the superior AUROC achieved by \emph{GT-GRN}. In particular, the mESC-1000 dataset highlights the clear dominance of \emph{GT-GRN}, with its AUROC surpassing that of all other methods by a substantial margin. 

\begin{table}[!htp]
    \centering
    \caption{\bf{AUROC and AUPRC scores for different methods across the various scRNA-seq datasets.}}
    \label{tab:auroc_auprc}
    \begin{adjustbox}{height = 1.4in, width=3.48in}

    \begin{tabular}{llll}
        \hline
       \rowcolor[HTML]{D6DBDF} \textbf{Dataset} & \textbf{Method}    & \textbf{AUROC} & \textbf{AUPRC} \\
        \hline
        \multirow{4}{*}{mESC-1000} 
                 & \emph{GT-GRN}   & \textbf{0.9483} & \textbf{0.8990} \\
                 & GNNLink  & 0.8833 & 0.8660 \\
                 & GENELink & 0.9133 & 0.8103 \\
                 & GNE      & 0.8984 & 0.8925 \\ 
        \hline
        \multirow{4}{*}{mESC-500} 
                 & \emph{GT-GRN}   & \textbf{0.9402} & \textbf{0.8853} \\
                 & GNNLink  & 0.8768 & 0.8331 \\
                 & GENELink & 0.9057 & 0.8004 \\
                 & GNE      & 0.8378 & 0.8416 \\
        \hline
        \multirow{4}{*}{hESC-500} 
                 & \emph{GT-GRN}   & \textbf{0.8793} & 0.5932 \\
                 & GNNLink  & 0.8251 & 0.4542 \\
                 & GENELink & 0.8618 & 0.5581 \\
                 & GNE      & 0.8402 & \textbf{0.8466} \\
        \hline
        \multirow{4}{*}{hESC-1000} 
                 & \emph{GT-GRN}   & 0.8784 & 0.8604 \\
                 & GNNLink  & 0.8442 & 0.5011 \\
                 & GENELink & 0.8657 & 0.5610 \\
                 & GNE      & \textbf{0.9025} & \textbf{0.9042} \\
        \hline

                 

    \end{tabular}
    \end{adjustbox}
\end{table}

 \begin{table}[h]   
\centering

\caption{\textbf{Tuned hyper-parameters for different models.} \textit{Epochs} sees the entire dataset during training, \textit{Learning Rate} controls the pace at which algorithm updates, \textit{Output Dimensions}  refers to the final layer's shape from which the predictions are made, \textit{Attention Heads} 
mechanism computes attention scores over the input elements and aggregates information. \textit{Layers} refer to the number of transformation steps that processes inputs and passes the transformed output to the next layer.}\label{tab: hyperparameters}
\begin{adjustbox}{height = 0.63in, width=3.40in}
\begin{tabular}{|l|l|l|l|l|l|}
\hline
\rowcolor[HTML]{D6DBDF}                                     
        \bf{Model}      & \bf{Epochs}    &  
        \begin{tabular}[c]{@{}c@{}} \bf{Learning} \\ \bf{Rate} \end{tabular} & \begin{tabular}[c]{@{}c@{}} \bf{Output} \\  \bf{Dimensions} \end{tabular}  & \begin{tabular}[c]{@{}c@{}} \bf{Attention} \\ \bf{Heads} \end{tabular} &  \bf{Layers}   \\  \cline{1-6}

\emph{GT-GRN} & - & 0.001, 0.003, 0.0005 & - & 2,4,8 & 4,6,8 \\

GNNLink & 100, 200, 300 & 0.001, 0.005, 0.01 & 128, 256, 512 & - & \\

GENELink & 10, 20, 30 & 0.001, 0.003, 0.0005 & 128, 256, 512 & - & - \\

GNE & 10, 20, 30& 0.001, 0.005, 0.01 & 128, 256, 512 & -& - \\
\hline

\end{tabular}
\end{adjustbox}
\end{table}

\section{Application of \emph{GT-GRN} on Cell-Type Classification}

Cell-type-specific gene regulatory networks (GRNs) are crucial for defining transcriptional states during development, with each cell-type being characterized by a unique set of active transcription factors (TFs). These GRNs offer an unbiased method for studying gene regulation, providing valuable insights into the mechanisms driving cellular diversity. In this context, we explore the effectiveness of \emph{GT-GRN} in reconstructing cell-type-specific GRNs, with the goal of cell-type annotation. To achieve this, we apply \emph{GT-GRN} to single-cell RNA sequencing (scRNA-seq) data from over 8,000 human peripheral blood mononuclear cells (PBMCs8k), sourced from 10X Genomics\footnote{\url{https://www.10xgenomics.com/datasets/8-k-pbm-cs-from-a-healthy-donor-2-standard-2-1-0}}. The data is preprocessed using the Scanpy framework\cite{wolf2018scanpy}, ensuring efficient handling and analysis of the single-cell data. For the ground truth network, we utilize the hTFtarget database\cite{zhang2020htftarget}, which integrates ChIP-seq data, transcription factor binding sites, and epigenetic modification information. This comprehensive resource provides detailed insights into gene regulation and TF-target interactions, making it an invaluable tool for studying gene regulatory mechanisms.

In order to perform cell-type annotation, it is essential to first reconstruct the GRN. This involves generating embeddings that represent the cell types based on their gene regulatory networks. These embeddings serve as the foundation for cell-type classification, enabling accurate annotation of the cell types based on their unique regulatory patterns. By leveraging the power of \emph{GT-GRN} in inferring cell-type-specific GRNs, we aim to advance the cell-type annotation in scRNA-seq data, ultimately improving our understanding of the complex regulatory landscapes that define cellular identities.

\subsection{\emph{GT-GRN} for PBMC Network Reconstruction}
We investigate GRN inference using \emph{GT-GRN} by formulating it as a network regeneration problem on the PBMC dataset. First, we filter the data by removing genes expressed in fewer than 5\% of cells and discarding cells that express fewer than 200 genes. Next, we normalize the total counts per cell to 10,000, ensuring comparability across cells. To further refine the data, we apply MAGIC\cite{van2018recovering} imputation, which reduces noise and improves expression patterns. Finally, we perform a logarithmic transformation to improve interpretability and optimize the data for downstream analysis.

After preprocessing, we employ \emph{GT-GRN} to reconstruct the PBMC gene regulatory network and compare its performance against existing baseline methods. Table \ref{tab:comparison} reveals the comparison of PBMC’s hTFTarget (Gold standard) and generated networks reveals key structural differences and predictive performance variations. \emph{GT-GRN} achieves the highest AUROC (0.9769) while maintaining balanced connectivity and clustering, making it the best-performing model. GENELink and GNNLink exhibit dense connectivity, high clustering, and shorter path length but are highly disassortative, indicating a strong preference for high-degree nodes connecting to low-degree ones. GNE, with lower connectivity and clustering, results in longer path lengths and the lowest AUROC (0.7596) but retains some structural similarities to the Gold network. The Gold standard itself maintains moderate connectivity and a sparse clustering structure, serving as a key benchmark. We also measure the quality of the generated graph network charactersitics from the candidate methods with the input network (hTFTarget) using a single measurement score using Pearson correlation coefficient. The results show that all methods exhibit strong correlation, with GNE (0.9992) achieving the highest agreement, followed closely by \emph{GT-GRN} (0.9838). However, GNNLink and GENELink report the similar score of 0.9817. 
These insights highlight the trade-offs between network structure and predictive performance, guiding model selection for biological network analysis.

\begin{figure}[h]
    \centering
    \includegraphics[width=8.7cm, height=6.2cm]{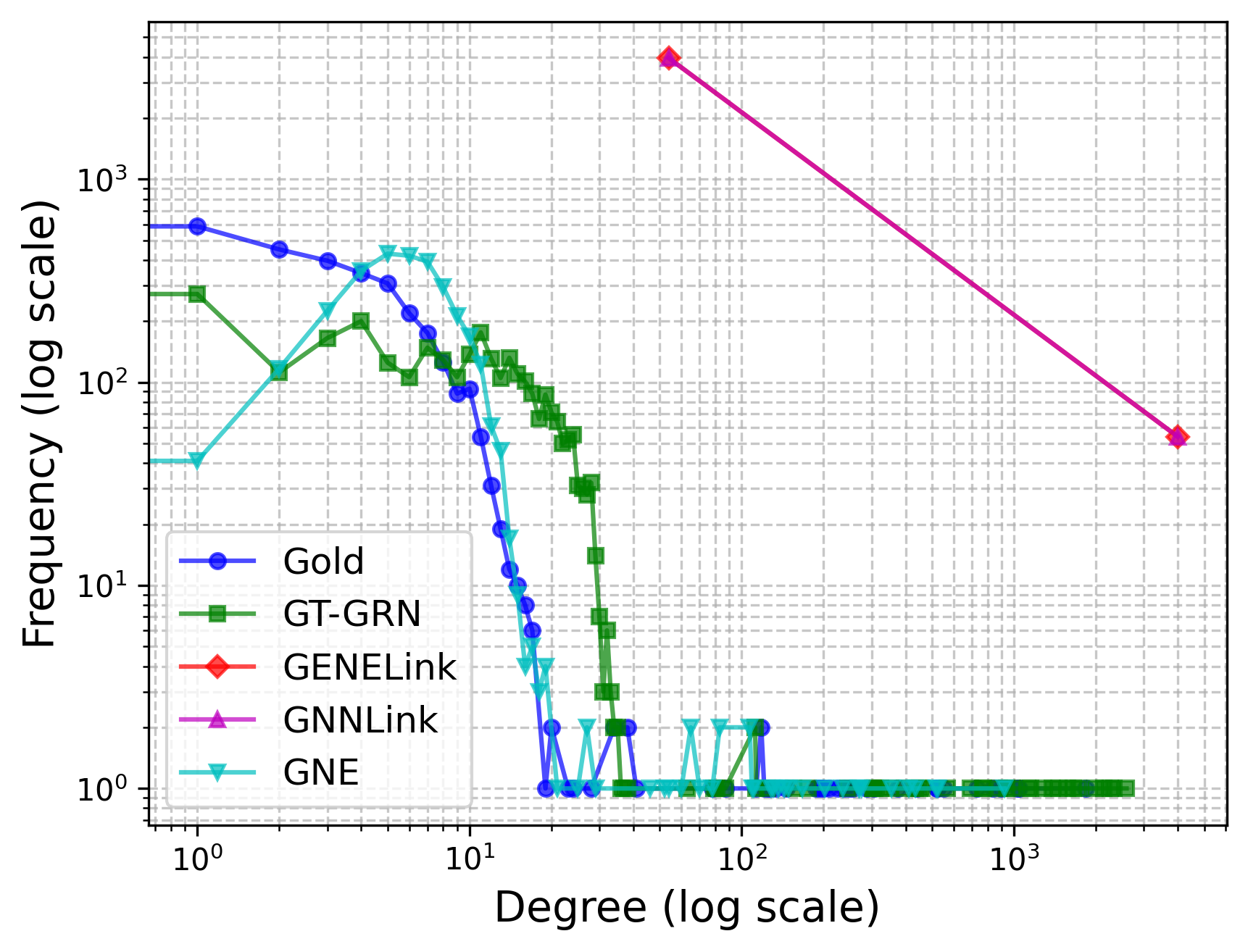}
    \caption{\textbf{Log-Log Degree Distribution plot of the input PBMC's hTFTarget Network and Generated Network for different models.}}
    \label{fig:degree_distribution}
\end{figure}

Further, we analyze the degree-distribution plot of the generated networks in comparison to the input PBMC's hTFTarget network. Figure \ref{fig:degree_distribution} describes the log-log degree distribution plot compares the degree distributions of the input (Gold) and generated networks (\emph{GT-GRN}, GENELink, GNNLink, and GNE). The Gold network follows a natural decay which is scale-free in degree distribution, while \emph{GT-GRN} shows a similar trend with slight deviations. GNE displays a more scattered pattern, indicating the similar tailed degree-distribution. GENELink and GNNLink exhibit significantly higher maximum degrees, i.e., these generated networks generate more high-degree nodes than the input network.

\begin{figure*}
    \centering
    \includegraphics[width=17cm, height=4.9cm]{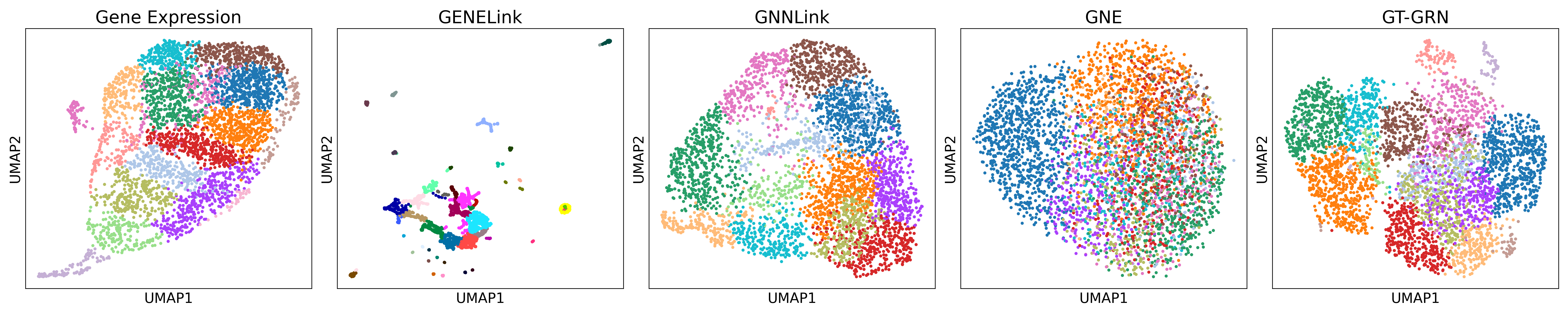}
    \caption{\textbf{UMAP visualization of genes representations for PBMC's dataset according to different methods.}} 
    \label{fig:umap}
\end{figure*}

Next, we assess how well our embeddings capture the community structure by clustering gene embeddings. We utilize the Leiden algorithm ~\cite{traag2019louvain} to cluster the resultant gene embeddings. Figure \ref{fig:umap} presents a uniform manifold approximation and projection (UMAP) dimensional reduction of gene representations for various methods. The methods compared include Gene Expression data, GENELink, GNNLink, GNE, and \emph{GT-GRN}. The visualization clearly shows the distinct community structures produced by GNNLink, and \emph{GT-GRN} embeddings that indicates effective preservation of biological modules. 

\begin{table}[h]
\centering
\caption{\textbf{Network Characteristics Comparison of PBMC's hTFTarget and Generated Networks with AUROC Score}. \emph{Maximum degree} computes the degree over all vertices. \emph{Assortativity} is the pearson correlation of degrees of  connected nodes. \emph{Triangle Count} denotes the connection between two nodes. \emph{Clustering Coefficient} measure of the tendency of nodes in a network to form triangles. \emph{Characteristic Path Length} represents the average shortest path length between all nodes pairs in a network. \emph{PCC} is the Pearson Correlation Coefficient between hTFTarget and generated network characterisitcs.}\label{tab:characteristics}
\begin{adjustbox}{height = 0.7in, width=3.5in}

\tabcolsep=0.20cm
\begin{tabular}{|l|l|l|l|l|l|l|l|l|}
\hline

\rowcolor[HTML]{D6DBDF}  \bf{Model} & 
\begin{tabular}[c]{@{}l@{}}\textbf{Maximum} \\ \textbf{Degree}\end{tabular} &
\begin{tabular}[c]{@{}l@{}}\textbf{Assort-} \\ \textbf{ativity}\end{tabular} &
\begin{tabular}[c]{@{}l@{}}\textbf{Triangle} \\ \textbf{Count}\end{tabular} &
\begin{tabular}[c]{@{}l@{}}\textbf{Clustering} \\ \textbf{Coefficient}\end{tabular} &
\begin{tabular}[c]{@{}l@{}}\textbf{Characteristic} \\ \textbf{Path Length}\end{tabular} & 
\begin{tabular}[c]{@{}l@{}}\textbf{AUROC}\end{tabular} & \bf{PCC} \\ \hline 

\rowcolor[HTML]{E8DAEF}

Gold & 1837 & -0.4762 & 9596 & 0.0060 & 2.7392 & - & - \\
 \emph{GT-GRN} & 2593 & -0.6614 & 207579 & 0.0255 & 2.1934 & 0.9769 & 0.9838\\
 GENELink & 3999 & -0.9867 & 5671530 & 0.0389 & 1.9731 & 0.8810 & 0.9817\\
 GNNLink & 3999 & -0.9867 & 5671530 & 0.0389 & 1.9731 & 0.8467 & 0.9817 \\
 GNE & 924 & -0.1872 & 4014 & 0.0094 & 2.9621 & 0.7596 & 0.9992\\ \hline 

\end{tabular}\label{tab:comparison}
\end{adjustbox}
\end{table}

\subsection{GT-GRN for Cell-Type Annotation}

\begin{figure}
    \centering
    \fbox{\includegraphics[width=8.6cm, height=5.8cm]{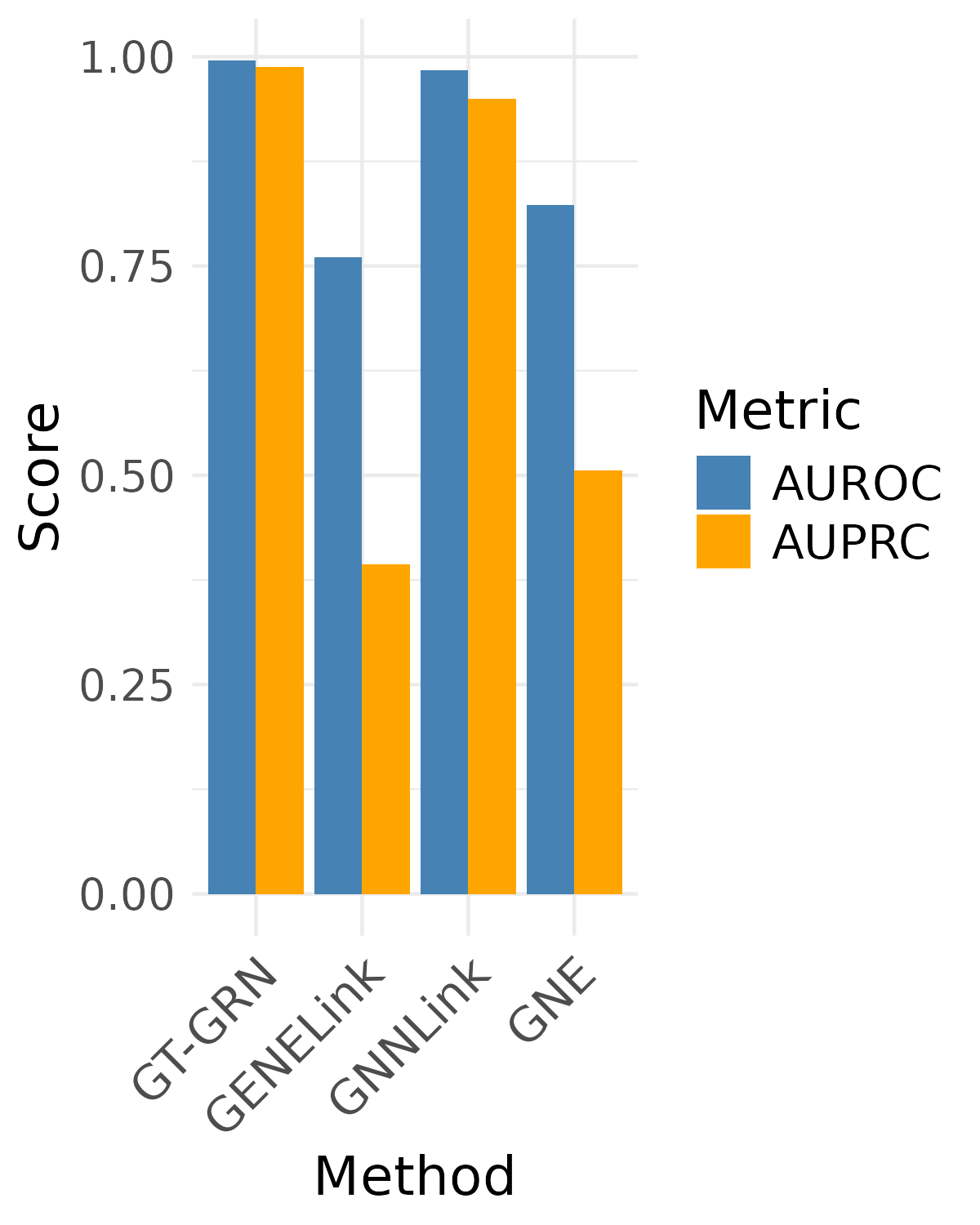}}
    \caption{\textbf{Cell-Type Classification}. \emph{AUROC and AUPRC score of GT-GRN, GENELink, GNNLink, GNE based embeddings in annotation cell-types using MLP classifier in 5 fold cross-validation
setting.}}
    \label{fig:cell_type}
\end{figure}

We further investigate these embeddings for cell-type annotation task. We manually annotate cell-types for gene expression data and focus on four cell-types with highest number of cells, \emph{CD4+ T} cells, \emph{CD14+ Monocyte cells}, \emph{CD8+ T cells}. We train a three-layered multi-layered preceptron classifier for annotating cell-types. The classifier is trained using in multi-class classifcation setting in five-fold cross validation setting. We benchmark \emph{GT-GRN} against GENELink, GNNLink and GNE methods. Figure \ref{fig:cell_type} demonstrate that \emph{GT-GRN} effectively captures the cell-types using gene representation classification setup.

Next, we delve into the individual contributions of each embedding modality within the \emph{GT-GRN} framework through a systematic ablation study on the PBMC dataset. This dataset provides a biologically rich and diverse single-cell expression landscape, making it an ideal benchmark for evaluating the role of each embedding component in GRN inference.

\subsection{Ablation Studies} To assess the overall efficacy and robustness of \emph{GT-GRN}, we performed a comprehensive ablation study that systematically examines the individual and combined contributions of its embedding modalities: structural positional encodings, global embeddings, and gene expression embeddings. This experiment is crucial, given the multi-component nature of our framework, as it allows us to systematically assess the role and efficacy of each component in the GRN inference process under a link prediction setup. We organize our baselines into three categories: uni-modal, bi-modal, and tri-modal, where the prefixes "uni", "bi", and "tri" denote the number of information sources used. This study establishes the necessity of each module in the overall architecture, demonstrating that omitting any single modality leads to a significant performance drop—thereby justifying the integration of all components for optimal GRN reconstruction.
\begin{itemize}
    \item \emph{Structural Positional Encodings} : In this uni-modal baseline, we have used the $PE_{graph}$ of the input network which is then fed into the \textit{GT-GRN}. Here, each gene positional encodings is of length 512. This vector is fed into the model to predict possibility of link with other gene vectors based on the input link information. 

    \vspace{0.2cm}

    \item \emph{Global Embeddings} : This is a uni-model baseline, that extracts the global knowledge of inferred GRNs in a multi-network integration framework using BERT model. The output length of each gene embedding here is 512. The vector is given as input into the model to estimate the likelihood of forming links with other gene vectors for link inference.

    \vspace{0.2cm}

    \item \emph{Gene Expression Embeddings} : In this uni-modal baseline, raw gene expression is converted into a embedding vector to capture the latent information using auto-encoder model. The length of the embedding vector for each gene is 512 which is then used to estimate the link likelihood using \emph{GT-GRN} framework. 

    \vspace{0.2cm}

    \item \emph{Structural Positional Encodings + Global Embeddings} : 
    This multi-modal approach combines the structural positional encodings of the input network with global embeddings derived from the BERT-based multi-network integration framework. The fused representation is used in the \textit{GT-GRN} model to enhance link prediction performance.

    \vspace{0.2cm}

    \item \emph{Structural Positional Encodings + 
    Gene Expression Embeddings} : This baseline integrates the structural positional encodings of the input network with gene expression embeddings obtained through an autoencoder. The combined vector representation is fed into the \textit{GT-GRN} model to infer potential links between genes.

    \vspace{0.2cm}

    \item \emph{Global Embeddings + 
    Gene Expression Embeddings}

    In this setup, global embeddings capturing inferred GRN knowledge are combined with gene expression embeddings. The resulting representation is used to predict gene interactions within the \textit{GT-GRN} framework.

    \vspace{0.2cm}

    \item \emph{Structural Positional Encodings + Global Embeddings + Gene Expression Embeddings} :
    This comprehensive multi-modal approach fuses all three embeddings—structural positional encodings, global embeddings, and gene expression embeddings—to provide a richer representation for link inference. This integrated approach aims to leverage complementary information from multiple modalities to enhance predictive performance.
\end{itemize}

Table \ref{tab:ablation} reports the ablation study results from different feature sets for \emph{GT-GRN} using AUROC scores. Among unimodal representations, Global Embeddings achieve the highest AUROC of 0.8860, followed by Gene Expression Embeddings 0.8693 and Structural Positional Encodings with AUROC score of 0.8480, indicating that global information is the most informative for link inference.  

Bimodal combinations improve performance, with Global Embeddings + Structural Positional Encodings 0.8843 and Global Embeddings + Gene Expression Embeddings 0.8841 performing best. The trimodal combination of all three embeddings achieves the highest AUROC 0.8877, demonstrating that integrating multiple modalities provides the most effective representation for GRN inference.

Overall, the study highlights the importance of multi-modal integration, with Global Embeddings playing a key role in enhancing predictive performance.

\begin{table}[h]
    \centering
    \caption{\textbf{Ablation study in terms of different features for \textit{PBMC's} dataset for \emph{GT-GRN} framework.}}\label{tab:ablation}
    \begin{adjustbox}{height = 1.6in, width=3.52in}
    \begin{tabular}{|l|l|c|}
        \hline
    \rowcolor[HTML]{D6DBDF}    \textbf{Modal} & \textbf{Feature sets} & \textbf{AUROC}  \\
        \hline
        Unimodal & Structural Positional Encodings  & 0.8480    \\
        \cline{2-3}
                 & Global Embeddings  & 0.8860    \\
        \cline{2-3}
                 & Gene Expression Embeddings  & 0.8693    \\
        \hline

        Bimodal  & \begin{tabular}[c]{@{}c@{}} Structural Positional Encodings \\ + \\ Global Embeddings \end{tabular}     & 0.8843      \\
        \cline{2-3}
                 & \begin{tabular}[c]{@{}c@{}} Structural Positional Encodings \\ + \\ Gene Expression Embeddings \end{tabular}    &  0.8666   \\
        \cline{2-3}
                 & \begin{tabular}[c]{@{}c@{}} Global Embeddings \\ + \\ Gene Expression Embeddings \end{tabular}  &  0.8841   \\
        \hline
        Trimodal & \begin{tabular}[c]{@{}c@{}} Structural Positional Encodings \\ + \\ Global Embeddings \\  +  \\ Gene Expression Embeddings \end{tabular} & \textbf{0.8877}     \\
        \hline
    \end{tabular}
    \end{adjustbox}
\end{table}

\section{Conclusion}
In this work, we proposed a novel gene regulatory network (GRN) inference framework, \emph{GT-GRN}, which leverages Graph Transformers to infer regulatory links by incorporating graph-based techniques. Our approach begins by generating gene embeddings through an autoencoder. We then integrate prior network knowledge from known GRNs using an NLP-based BERT model, where these graphs are converted into sequences to extract contextual embeddings. Additionally, we incorporate graph positional information to enhance the inference process.

Through extensive experiments, \emph{GT-GRN} demonstrates superior link prediction performance compared to baseline methods. We further assess the quality of the generated embeddings by evaluating their community structure, showing that \emph{GT-GRN} effectively supports cell-type annotation in real PBMC gene expression datasets. Our ablation study reveals that the combination of gene expression data, global gene context, and positional information significantly contributes to improved GRN inference precision. While the current work focuses on accurate reconstruction of Gene Regulatory Networks (GRNs) using multi-modal embeddings, ongoing efforts are directed toward extending the framework to prioritize disease-associated genes. By leveraging the inferred GRNs, the goal is to identify key regulatory hubs and pathways that may play pivotal roles in disease development and progression.

\section{Acknowledgement}
This work is supported by IDEAS-TIH, ISI-Kolkata research project (funded by Department of Science \& Technology (DST), Govt. of India) at the Network Reconstruction and Analysis (NETRA) Lab, Department of Computer Applications, Sikkim (Central) University.

\bibliographystyle{unsrt}{}
\bibliography{references}{}

\end{document}